\documentclass[dvipsnames,nonacm,format=sigconf,anonymous=false,review=false]{acmart}

\usepackage[ruled,linesnumbered]{algorithm2e}
\usepackage{amsmath}   
\usepackage{booktabs}
\usepackage{tabularx}
\usepackage{multirow}
\usepackage{siunitx} 
\usepackage{graphicx}
\usepackage{bm}
\usepackage{subcaption}
\usepackage{url}

\AtBeginDocument{%
  }

\setcopyright{acmlicensed}
\copyrightyear{2018}
\acmYear{2018}
\acmDOI{XXXXXXX.XXXXXXX}
\acmConference[Conference acronym 'XX]{Make sure to enter the correct
  conference title from your rights confirmation email}{June 03--05,
  2018}{Woodstock, NY}
\acmISBN{978-1-4503-XXXX-X/2018/06}

\begin{document}

\title{Surrogate Ensemble in Expensive Multi-Objective \\Optimization via Deep Q-Learning}


\author{Yuxin Wu}
\authornote{Both authors contributed equally to this research.}
\email{wuyuxin418@gmail.com}
\orcid{0009-0002-0277-1311}
\affiliation{%
  \institution{South China University of Technology}
  \city{Guangzhou}
  \state{Guangdong}
  \country{China}
}

\author{Hongshu Guo}
\authornotemark[1]
\email{guohongshu369@gmail.com}
\orcid{0000-0001-8063-8984}
\affiliation{%
  \institution{South China University of Technology}
  \city{Guangzhou}
  \state{Guangdong}
  \country{China}
}

\author{Ting Huang}
\email{gnauhgnith@gmail.com}
\orcid{0000-0002-8755-043X}
\affiliation{%
  \institution{Xidian University}
  \city{Guangzhou}
  \state{Guangdong}
  \country{China}
}

\author{Yue-Jiao Gong}
\email{gongyuejiao@gmail.com}
\orcid{0000-0002-5648-1160}
\affiliation{%
  \institution{South China University of Technology}
  \city{Guangzhou}
  \state{Guangdong}
  \country{China}
}

\author{Zeyuan Ma}
\email{scut.crazynicolas@gmail.com}
\orcid{0000-0001-6216-9379}
\affiliation{%
  \institution{South China University of Technology}
  \city{Guangzhou}
  \state{Guangdong}
  \country{China}
}
\authornote{Corresponding Author.}






\renewcommand{\shortauthors}{Trovato et al.}

\begin{abstract}
Surrogate-assisted Evolutionary Algorithms~(SAEAs) have shown promising robustness in solving expensive optimization problems. A key aspect that impacts SAEAs' effectiveness is surrogate model selection, which in existing works is predominantly decided by human developer. Such human-made design choice introduces strong bias into SAEAs and may hurt their expected performance on out-of-scope tasks. In this paper, we propose a reinforcement learning-assisted ensemble framework, termed as SEEMOO, which is capable of scheduling different surrogate models within a single optimization process, hence boosting the overall optimization performance in a cooperative paradigm. Specifically, we focus on expensive multi-objective optimization problems, where multiple objective functions shape a compositional landscape and hence challenge surrogate selection. SEEMOO comprises following core designs: 1) A pre-collected model pool that maintains different surrogate models; 2) An attention-based state-extractor supports universal optimization state representation of problems with varied objective numbers; 3) a deep Q-network serves as dynamic surrogate selector: Given the optimization state, it selects desired surrogate model for current-step evaluation. SEEMOO is trained to maximize the overall optimization performance under a training problem distribution. Extensive benchmark results demonstrate SEEMOO's surrogate ensemble paradigm  boosts the optimization performance of single-surrogate baselines. Further ablation studies underscore the importance of SEEMOO's design components.      
\end{abstract}




\maketitle

\section{Introduction}

Multi-objective optimization (MOO) is pervasive in complex real-world decision-making scenarios, such as integrated engineering systems~\cite{wang2015modelling} and neural architecture search~\cite{lu2019nsga}. Therefore, Evolutionary multi-objective optimization algorithms (MOEAs) have been widely adopted to address such problems due to their parallel search mechanism based on population, which effectively balances convergence and diversity to approximate the pareto front and obtain a diverse set of high-quality non-dominated solutions. Representative methods, including NSGA-II~\cite{deb2002fast} and MOEA/D~\cite{zhang2007moea}, have demonstrated remarkable performance on benchmark~\cite{zitzler2000comparison,huband2006review,tan2002evolutionary} and practical applications~\cite{reed2013evolutionary,li2015many,lin2019pareto}. However, as industrial problems grow increasingly complex, many real-world MOO tasks rely on expensive simulations or physical experiments, rendering objective function evaluations prohibitively costly and significantly limiting the practical applicability of conventional MOEAs.

To mitigate the high cost of evaluation, surrogate-assisted evolutionary algorithms (SAEAs)\cite{jin2018data} have emerged as one of the mainstream research directions. By introducing lightweight surrogate models—such as Gaussian Processes (GPs)~\cite{tian2021balancing} and Radial Basis Functions (RBFs)~\cite{guo2022edge}, to approximate the true objective functions, SAEAs have been successfully applied to a wide range of critical domains, including aerodynamic shape optimization in automotive crashworthiness design~\cite{hamza2005vehicle} and aerospace engineering~\cite{han2020recent}.

Despite their success, the effectiveness of surrogate-assisted evolutionary algorithms (SAEAs) is highly sensitive to the surrogate management strategy. Existing methods mainly rely on manually predefined surrogate models~\cite{jin2018data} or ensemble schemes~\cite{guo2018heterogeneous}, which inevitably introduce designer bias across diverse problems, thereby impairing both generalizability and robustness on unseen tasks. Recently, Meta‑Black‑Box Optimization (MetaBBO)~\cite{ma2025toward,yang2025meta} has emerged as a learning paradigm that integrates reinforcement learning (RL) and other machine learning techniques with evolutionary computation to reduce dependence on handcrafted algorithmic configurations. Notably, MetaBBO has achieved large success in single-objective optimization~\cite{ma2024auto,li2025b2opt,han2025enhancing}, differential optimization~\cite{guo2024deep}, multitask optimization~\cite{guo2025configx}, and multimodal optimization~\cite{lian2024rlemmo}, showing strong adaptability and generalization. However, integrating meta-learning into surrogate modeling for expensive multi-objective optimizations (EMOOs)~\cite{deng2025evolutionary} remains largely underexplored, despite these advances.

To bridge this gap, in this paper, we propose SEEMOO, a RL-assisted surrogate ensemble framework for EMOOs. SEEMOO incorporates a learning-driven surrogate scheduling mechanism into the optimization process, enabling the dynamic selection of suitable surrogate models to improve optimization efficiency. Specifically, SEEMOO first constructs a candidate pool of diverse surrogate models covering different modeling biases and application scenarios. Then, An attention-based state extractor is employed to unify the representation of population convergence, performance metrics, and multi-objective information, supporting optimization problems with varying numbers of objectives. Based on the extracted state, a Deep Q-Network (DQN) is trained to learn a dynamic policy that schedules surrogate models in response to the current optimization state.

To evaluate the proposed mechanism, we integrate SEEMOO into the NSGA-II framework and conduct experiments on representative benchmark problems. Results demonstrate that our dynamic scheduling strategy significantly outperforms various individual static surrogate models in optimization efficiency under limited evaluation budgets. Further ablation studies verify the effectiveness of the key components in SEEMOO, including the reward design and
the cooperative use of multiple surrogate models.

The main contributions of our work are summarized as follows:
\begin{itemize}
    \item We propose SEEMOO, a RL-assisted framework for expensive multi-objective optimization, enabling dynamic surrogate model scheduling under limited evaluation budgets.
    \item SEEMOO combines a diverse surrogate model pool, a state extractor based on attention mechanism for effective optimization state representation, and a DQN-based selector to adaptively choose the best model at each step.
    \item Through comparative studies and detailed ablation analyses within the NSGA-II framework, we verify that SEEMOO outperforms various individual static surrogate models and demonstrate the effectiveness of its core components.
\end{itemize}

\section{Related Works}
\subsection{Surrogate-Assisted Evolutionary Algorithms}
In expensive multi-objective optimization, evaluating functions often requires costly simulations or experiments. To reduce this cost, SAEAs employ lightweight approximate models as substitutes. Recent surveys provide comprehensive overviews of this field~\cite{he2023review,alizadeh2020managing}.

Early SAEA researches focused primarily on single-surrogate models such as Gaussian Processes (GPs)~\cite{zhou2005study}, Radial Basis Functions (RBFs)~\cite{regis2013combining}, K-Nearest Neighbors (KNN)~\cite{he2007hybridisation}, Multi-Layer Perceptron (MLP)~\cite{jin2005comprehensive} and Support Vector Regression (SVR)~\cite{ciccazzo2015svm}, ofen combined withe infill criteria to balance exploration and exploitation~\cite{jin2011surrogate,chugh2017surrogate}. However, single models often exhibit inherent biases and limited adaptability, prompting the development of ensemble methods~\cite{liu2025expensive} that combine multiple models for better performance. For instance, Goel et al.~\cite{goel2007ensemble} focused on constructing diverse models to offset individual biases, while Li et al.~\cite{li2020data} concentrated on leveraging predictive uncertainty for adaptive model weighting.

Though ensemble methods enhance surrogate robustness, their application was initially confined to single-objective optimization. Adapting these methods to the multi-objective domain requires addressing additional complexities: high-dimensional landscapes are tackled through variable grouping and landscape sampling~\cite{zhen2025surrogate, huang2025surrogate, liu2025variable, gu2025large}, expensive constraints are managed via adaptive or reinforcement learning-assisted strategies~\cite{liu2025fast, shao2025deep, gu2025exploring}, and heterogeneous objective structures are handled through specialized correlation estimation~\cite{gu2024local}. Recent works have also extended MOO with various technical innovations~\cite{si2023linear,zhang2023multigranularity,yang2023surrogate}, such as a information transfer mechanisms introduced by Luo et al.~\cite{luo2022expensive} to improve surrogate efficiency across objectives.

Despite these advances, a common limitation remains: most existing methods rely on static surrogate selection or combination strategies, which cannot adapt to the evolving optimization state. Motivated by this observation, we propose SEEMOO for dynamic surrogate selection, providing a more adaptive approach to EMOOs.

\subsection{Meta-Black-Box Optimization}
Recently, Meta-Black-Box Optimization (MetaBBO) has emerged as a learning-enhanced paradigm that integrates machine learning techniques with evolutionary computation for black-box optimization. It employs a bi-level framework where a meta-policy dynamically configures the base optimizer, reducing manual tuning and enhancing generalization. Relative surveys~\cite{ma2025toward} have laid a systematic foundation for research in this direction.

Within this paradigm, MetaBBO encompasses diverse learning methods, including RL (MetaBBO-RL)~\cite{ma2024auto,guo2025designx}, supervised learning (MetaBBO-SL)~\cite{chen2017learning,li2024pretrained}, neuroevolution (MetaBBO-NE)~\cite{lange2023discovering}, and in-context learning (MetaBBO-ICL)~\cite{ma2026llamoco,wu2023large} with systematic analyses in recent survey~\cite{ma2025toward,ma2023metabox,ma2025metabox}. Among these, RL has demonstrated remarkable progress in dynamic algorithm design due to its natural suitability for decision-making. Considering concrete algorithm design tasks, existing MetaBBO-RL works can be categorized into: 1) \emph{Algorithm Selection}: where a RL agent is used at the meta-level to dynamically select desired operator~\cite{r2rlmoea} or entire algorithm~\cite{hhrlmar,guo2024deep}, for the low-level optimization process, achieving flexible and enhanced optimization performance; 2) \emph{Algorithm Configuration}: where the parameters and/or operators of a low-level BBO algorithm are adjusted by the meta-level policy in response to the problem landscape, enabling fine-tuned optimization performance~\cite{guo2025reinforcement,guo2025configx,deddqn,ma2024auto,chen2025metade,sun2021learning,tan2021differential}; 3) \emph{Algorithm Generation}: where the meta-level policy is trained to manipulate and evolve solutions by applying evolutionary or heuristic operations, acting as a BBO algorithm~\cite{zhao2024automated,chen2024symbol}; 4) \emph{Solution Manipulation}: where the meta-level policy constructs novel algorithmic components and assembles their overall workflow, thereby generating entirely new algorithms \cite{chaybouti2022meta}.

While MetaBBO-RL has proven effective in the aforementioned categories, applying these methods to expensive multi-objective problems remains challenging. Although initial explorations have begun to combine learning with surrogates~\cite{ma2025surrogate}, most existing studies employ RL to construct high-level strategies or control components under constrained evaluation budgets. For example, Shao et al.~\cite{shao2025deep} explore surrogate management strategies, and Du et al.~\cite{du2025meta} develop a surrogate-assisted framework with adaptive operator tuning. However, a unified dynamic surrogate scheduling framework for expensive multi-objective optimization remains unexplored. We investigate this direction with SEEMOO, a meta-RL framework under the MetaBBO paradigm.
\section{Methodology}
\begin{figure}[t]
    \centering
    \includegraphics[width=\columnwidth]{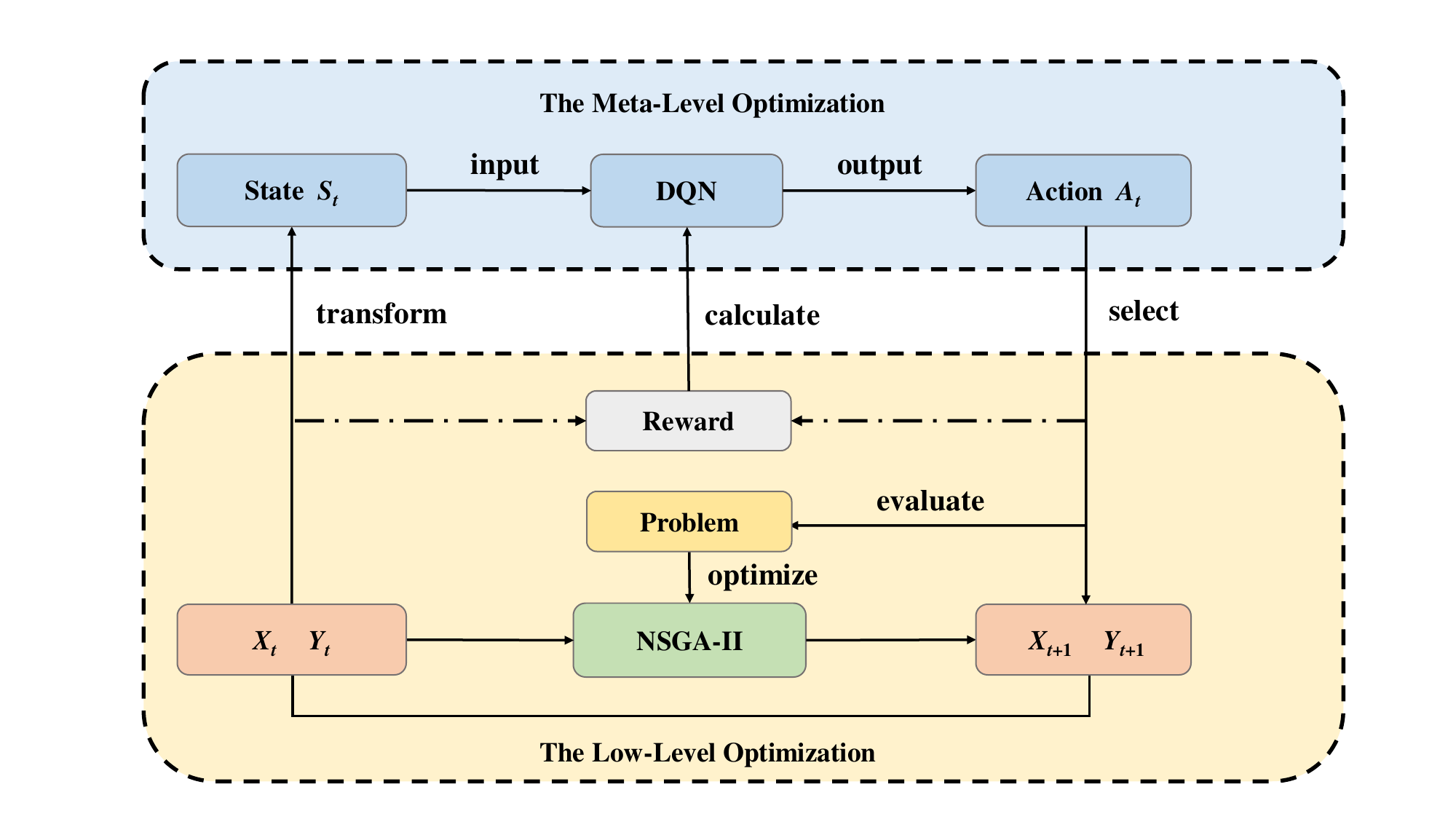} 
    \caption{Overall workflow of the proposed SEEMOO framework. The optimization process is formulated as a bi-level structure, consisting of a meta-level reinforcement learning controller and a low-level evolutionary optimizer.} 
    \Description{A block diagram illustrating the bi-level structure. The upper level shows a Meta-level Agent using DQN to select surrogate models based on state features. The lower level depicts the optimization process where the selected model guides NSGA-II, evaluates candidates, and updates the training database, finally returning a reward to the meta-level.}
    \label{fig:bi-level}
\end{figure}
\subsection{Overview}
As we mentioned before, surrogate model's performance significantly impacts the final performance of MOEAs when dealing with EMOOs. Motivated by this, we propose SEEMOO to ensemble capabilities of diverse surrogate models within MOEAs to further enhance the optimization performance. Specifically, we use a RL agent to dynamically schedule surrogate selection in NSGA-II during the optimization process. At each optimization step, the state feature comprising the surrogate performance and optimization progress are extracted and fed to the RL agent. The RL agent selects a surrogate model from a pre-defined model pool accordingly. The NSGA-II uses the selected surrogate model for next step optimization, a performance feedback (reward) is computed for training the RL agent. We illustrate the overall workflow of our SEEMOO in Figure~\ref{fig:bi-level}, and we detail each specific design in our framework in the following sections.

\subsection{Low-level Optimizer}
The low-level optimizer in SEEMOO is a surrogate-assisted NSGA-II~\cite{lu2019nsga}. Given a surrogate model $M$ from the pool $\mathcal{M}$, the optimization process starts with an initialization phase, where an initial population $X_{t=0} \in \mathbb{R}^{N\times dim}$ is generated via Latin hypercube sampling (LHS)~\cite{mckay2000comparison}, with $N$ denoting the population size and $dim$ the problem dimension.
Following the initialization, the optimization proceeds through three modular phases:
 
1) \emph{Initial Construction}: 
Each candidate is evaluated with the expensive true objective functions $F(x)$. The samples are stored in a training database $\mathcal{T} = \{(x, F(x))\}$ while an external archive $\mathcal{A}$ is initialized with the non-dominated solutions from $\mathcal{T}$ to track the global pareto front. 

2) \emph{Population Evolution}: At each optimization step $t$, the surrogate $M$ is trained on $\mathcal{T}$ to provide predictions $\hat{F}$. The current population $X_t$ then undergoes tournament selection, simulated binary crossover, and polynomial mutation guided by $\hat{F}$. The combined parent and offspring sets are ranked via non-dominated sorting and crowding distance based on $\hat{F}$, and the top $N$ individuals form the next generation $X_{t+1}$, without the true expensive evaluations.

3) \emph{Infilling and Update}: Following the surrogate-assisted search, A batch of $b$ promising candidates $X_{\mathrm{opt}} \subset X_{t+1}$ is selected from the first non-dominated front for true evaluation $F(x)$. The resulting pairs $(X_{\mathrm{opt}}, F(X_{\mathrm{opt}}))$ are appended to $\mathcal{T}$ to refine the surrogate and are used to update archive $\mathcal{A}$, retaining only global non-dominated solutions. Then, the optimization process continues iteratively until the maximum function evaluations $maxFEs$ are used up. Finally, the archive $\mathcal{A}$ is returned as the final approximated pareto front.

\subsection{Meta-level RL designs}
In SEEMOO, we first augment the single surrogate setting in the optimizer above towards dynamic surrogate selection for a ensemble and enhanced optimization performance. To learn such dynamic selection policy, we propose following RL designs. 
\subsubsection{State Space}
To provide the meta-policy with a comprehensive perception of the optimization status, we design a multidimensional state vector ${s}_t \in \mathbb{R}^{m \times 8}$, where $t$ denotes the current surrogate scheduling step, and $m$ is the number of candidate models. For each candidate model $\mathcal{M}_i \in \mathcal{M}, i\in\{0, \dots, m-1\}$, an 8-dimensional feature vector is extracted to characterize its selection history related to action $a_t$, IGD-based performance improvement, search convergence dynamics and progress.
\begin{enumerate}
   
    \item $s_{i,1}$ (Cumulative Usage). Represent the long-term preference for model $\mathcal{M}_i$:
    \begin{equation}
         s_{i,1} = \frac{C_{i,t}}{t},
    \end{equation}
   
    where $C_{i,t}$ denotes the total number of times model $\mathcal{M}_i$ has been selected up to step $t$.
    
    \item $s_{i,2}$ (Recent Frequency). The selection rate of model $\mathcal{M}_i$ over previous 5 steps to reflect recent policy tendencies:
    \begin{equation}
        s_{i,2} = \frac{1}{5} \sum_{j=1}^{5} G(a_{t-j} = i),
    \end{equation}
    where $G(\cdot)$ equals 1 if $a_{t-j} = i$ else 0.

    \item $s_{i,3}$ (Model-specific Improvement). The performance gain achieved by the model's most recent execution:
    \begin{equation}
         s_{i,3} = \frac{IGD_{i, j-1} - IGD_{i, j}}{IGD_{i,max} - IGD_{i,min}},
    \end{equation}
    where $j$ and $j-1$ denote the indices of the current and previous activation in the history of model $\mathcal{M}_i$. $IGD_{i,max}$ and $IGD_{i,min}$ denote the maximum and minimum IGD value achieved by model $\mathcal{M}_i$ in all its activation up to step $t$.
    \item $s_{i,4}$ (Global Step-wise Improvement). The global improvement in the last decision step:
    \begin{equation}
        s_{i,4} = \frac{IGD_{t-1} - IGD_t}{IGD_{max} - IGD_{min}},
    \end{equation}
    where $IGD_{t}$ and $IGD_{t-1}$ denote the IGD values at the current and previous optimization steps. $IGD_{max}$ and $IGD_{min}$ denote the maximum and minimum IGD value observed globally across all decision steps up to step $t$.

    \item $s_{i,5}$ (Decision Space Convergence) . The rate of change in population density within the $X$ space:
    \begin{equation}
         s_{i,5} = \frac{D_{X, t-1} - D_{X, t}}{D_{X,max} - D_{X,min}},
    \end{equation}
   
    where $D_X$ denotes the mean Euclidean distance between individuals in the decision space. $D_{X,max}$ and $D_{X,min}$ denote the maximum and minimum values of decision space density change up to step $t$.
     
    \item $s_{i,6}$ (Objective Space Convergence). The rate of change in population density within the $F$ space:
    \begin{equation}
        s_{i,6} = \frac{D_{F, t-1} - D_{F, t}}{D_{F,max} - D_{F,min}},
    \end{equation}
    where $D_F$ denotes the mean distance in the objective space. $D_{F,max}$ and $D_{F,min}$ denote the maximum and minimum values of objective space density change up to step $t$.

    \item $s_{i,7}$ (Surrogate Error Variation). The improvement in approximation accuracy on the training base $\mathcal{T}$:
    \begin{equation}
         s_{i,7} = \frac{E_{i, t-1} - E_{i, t}}{E_{i,max} - E_{i,min}},
    \end{equation}
    where $E_i$ denotes the Mean Squared Error (MSE) of model $\mathcal{M}_i$. $E_{i,max}$ and $E_{i,min}$ denote the maximum and minimum MSE values of model $\mathcal{M}_i$ up to step $t$.
    \item $s_{i,8}$ (Step Ratio). A feature representing the search progress:
    \begin{equation}
          s_{i,8} = \frac{t}{t_{max}},
    \end{equation}
    where $t_{max}$ denotes the total number of optimization steps.

\end{enumerate}

\subsubsection{Action Space}
We define the action space as a selection from a pool of five distinct surrogate models:
\begin{itemize}
    \item $a_0$: GP~\cite{zhou2005study};
    \item $a_1$: MLP1~\cite{jin2005comprehensive} with low budget;
    \item $a_2$: MLP2~\cite{jin2005comprehensive} with high budget;
    \item $a_3$: KNN~\cite{he2007hybridisation};
    \item $a_4$: RBFN~\cite{regis2013combining};
\end{itemize}
For an easier implementation, we use the \emph{sklearn}\footnote{\url{https://scikit-learn.org/stable/user_guide.html}} toolbox. We use the default hyper-parameters defined in \emph{sklearn} for these models expect few modifications: 1) for GP, using a kernel $C(1.0) \times RBF(1.0)$ with length-scale bounds $(10^{-2}, 10^{2})$ with optimization restarts ($2$ times); 2) for KNN, setting the neighborhood size to 5; 3) for both MLPs, using two hidden layers with $32$ units each, but limiting training iterations to $300$ for the low-budget version and $500$ for the high-budget version; 4) for RBFN, using $50$ radial basis centers determined by K-means clustering.

\subsubsection{Reward Function}
We define the reward $r_t$ based on the improvement of IGD of the external archive $\mathcal{A}$ between consecutive steps. To ensure the reward is well-scaled across different optimization problems, we normalize the improvement as shown in Eq.(4):
\begin{equation}
    r_t = \frac{IGD_{t-1} - IGD_t}{IGD_{max} - IGD_{min}}.
\end{equation}

If $IGD_t < IGD_{t-1}$, the agent receives a positive reward, encouraging actions that lead to a better approximation of the pareto front, otherwise the agent receives a negative reward.

\subsubsection{Network Design}
To effectively process the optimization features and derive the optimal surrogate scheduling policy, SEEMOO
adopts a Deep Q-Network (DQN) agent, denoted as $\pi_\theta$ parameterized by $\theta$, which adaptively selects the most appropriate surrogate model by mapping the state matrix $s_t$ to action values. The network architecture comprises two components: a feature extraction module ($\theta_{\mathrm{feat}}$) and a policy decision module ($\theta_{\mathrm{act}}$).

\paragraph{Feature Extraction Module} This module ($\theta_{\mathrm{feat}}$) transforms the raw features through a sequence of linear and relational operations. The dimension transformation follows:
\begin{equation} 
    s_t \xrightarrow{\text{Linear}} \mathbb{R}^{m \times 32} \xrightarrow{\text{Attention}} \mathbb{R}^{m \times 32} \xrightarrow{\text{LayerNorm}} H_{\mathrm{feat}},
\end{equation}
where $H_{\mathrm{feat}} \in \mathbb{R}^{m \times 32}$ is the output feature matrix. Each 8-D surrogate feature is first projected to 32-D via a shared linear layer, then processed by a 4-head Multi-Head Attention~\cite{vaswani2017attention} to capture inter-surrogate relationships, followed by Layer Normalization for training stability.

\paragraph{Policy Decision Module}
The decision module maps the embeddings into the final state-action value space. Let $h_\mathrm{feat}^i$ denotes the $i$-th row of $H_\mathrm{feat}$, corresponding to the embedded features of surrogate $\mathcal{M}_i$. Each $h_\mathrm{feat}^i$ is processed through a shared linear layer with parameters $\theta_\mathrm{act}$, followed by ReLU activation, yielding the Q-value:
\begin{equation}
Q(s_t, a_i) = \text{ReLU}\bigl(\text{Linear}(h_\mathrm{feat}^{i}; \theta_{\mathrm{act}})\bigr), \quad i = 0, \dots, m-1.
\end{equation}
This operation transforms the $m \times 32$ feature matrix into an $m$-dimensional vector $Q$, where each element $Q(s_t, a_i)$ estimates the expected value of selecting the model $\mathcal{M}_i$.

\subsection{Overall Workflow}
The overall workflow of SEEMOO follows a bi‑level meta‑learning paradigm, where a meta‑level RL agent dynamically schedules surrogate models for the low‑level NSGA-II optimizer. And the interaction is learned through a training process, which is detailed in Algorithm~\ref{alg:seemoo-training}: 

1) \emph{Initialization (Lines 4--8)}: The training begins by initializing the meta-policy $\pi_\theta$ and an experience replay buffer $\mathcal{R}$. The framework iterates through $N_{\mathrm{epoch}}$ epochs across a problem distribution $\mathcal{I}$. For each problem instance $P \in \mathcal{I}$, an initial population $X_{t=0}$ is generated via Latin Hypercube Sampling (LHS) and evaluated using the expensive true objectives $F(x)$. These samples construct the initial training database $\mathcal{T}$ and extract the starting state $s_{t=0}$, while the evaluation counter $FE$ is initialized to track the consumption of the evaluation budget.

2) \emph{Optimization Loop (Lines 10--13)}: At the core of training phase is an iterative interaction loop that persists until the evaluation budget $maxFEs$ is exhausted. At each step $t$, the agent observes the current state $s_t$ and selects an action $a_t$ via an $\epsilon$-greedy policy, which identifies the selected surrogate model $M$ from the pool $\mathcal{M}$. After training $M$ on the latest database $\mathcal{T}$, the surrogate-assisted NSGA-II is invoked. The population $X_t$ evolves through surrogate-approximated search to identify a batch of promising candidates $X_{\mathrm{opt}}$, which are evaluated with the true evaluation $F(x)$ to refine the surrogate model.

3) \emph{Policy Updating (Line 14--19)}:  Upon obtaining real evaluations, the database $\mathcal{T}$ is updated with new samples $(X_{\mathrm{opt}}, F(X_{\mathrm{opt}}))$. The framework then computes the IGD-based reward $r_t$ and extracts the next state $s_{t+1}$ to perceive the search progress (Lines 15–16) . And the transition $(s_t, a_t, r_t, s_{t+1})$ is stored in the buffer $\mathcal{R}$. The policy $\pi_\theta$ is then updated through DQN , which samples transitions from the buffer and optimizes the network parameters. The counters are updated, then the loop iterates until the evaluation budget $maxFEs$ is exhausted, finally returning the trained policy $\pi_\theta$ after all epochs are processed.

In summary, SEEMOO's workflow forms a full cycle: the meta-RL agent dynamically schedules surrogates based on real-time state, guiding the low-level evolutionary search; Then the resulting optimization feedback refines the agent's policy. This design not only automates the traditionally manual surrogate‑scheduling process, but also ensures that the policy improves iteratively through interaction with the optimizer.
\begin{algorithm}[t]
\caption{The SEEMOO Training Framework}
\label{alg:seemoo-training}
\SetKwInOut{Input}{Input}\SetKwInOut{Output}{Output}

\Input{Problem distribution $\mathcal{I}$, maximum function evaluations $maxFEs$, surrogate pool $\mathcal{M}$, maximum training epochs $N_{\mathrm{epoch}}$}
\Output{Optimal meta-policy $\pi_\theta$}

Initialize $\pi_\theta$ and experience buffer $\mathcal{R}$\;
\BlankLine
\For{epoch $ \leftarrow 0$ \KwTo $N_{\text{epoch}}$}{
    \For{problem $P$ in $\mathcal{I}$}{
        $t \leftarrow 0$ \;
        Initialize $X_{t=0}$ via LHS and evaluate with $F(x)$\;
        Extract initial state $s_{t=0}$ following Section 3.3.2\; 
        $\mathcal{T} \leftarrow \{(x, F(x)) \mid x \in X_{t=0}\}$\;
         $FE \leftarrow |X_{t=0}|$\;
        \While{$FE < maxFEs$}{
            $a_t \leftarrow \pi_\theta(s_t)$ with $\epsilon$-greedy policy\;
            $M \leftarrow \mathcal{M}(a_t)$ \;
            Train $M$ on $\mathcal{T}$\;
            $(X_{t+1}, X_{\mathrm{opt}}) \leftarrow \text{NSGA-II}(M, \mathcal{T})$ following Section 3.2 \;           
            $\mathcal{T} \leftarrow \mathcal{T} \cup \{(X_{\mathrm{opt}},F_{\mathrm{opt}})\}$\; 
            Compute reward $r_t$ following Section 3.3.3\;
            Extract next state $s_{t+1}$ following Section 3.3.2\;      
            $\mathcal{R} \leftarrow \mathcal{R} \cup \{(s_t, a_t, r_t, s_{t+1})\}$\;
            Update policy $\pi_\theta$ following Section 3.3.4\;
            $t \leftarrow t + 1$, $FE \leftarrow FE + |X_{\mathrm{opt}}|$\;         
        }
    }
}
\Return{$\pi_\theta$}\;
\end{algorithm}

\section{Experiments}
\subsection{Experimental Setup}
\subsubsection{Optimization Problem Configurations}
We evaluate the proposed framework on a benchmark suite of 24 multi-objective optimization problems implemented in the \emph{pymoo} framework~\cite{blank2020pymoo}. The suite covers the ZDT~\cite{zitzler2000comparison}, DTLZ~\cite{deb2002scalable}, and WFG~\cite{huband2006review} families, as well as the OmniTest~\cite{huband2006review} and Kursawe~\cite{kursawe1990variant} problems, thereby encompassing a wide range of pareto front characteristics and landscape complexities. As summarized in Table~\ref{tab:problem_dims}, the problems span decision variable dimensions $n_{var}$ from 2 to 80 and objective dimensions $n_{obj}$ up to 5. For problem instances without explicit parameter specifications, the default \emph{pymoo} configurations are adopted.

\subsubsection{Experimental Partitioning Design}
To assess cross-problem generalization, we construct two complementary training--testing configurations with different data availability regimes. Let $\mathcal{I}_{all}=\{1,2,\dots,24\}$ denote the full problem set listed in Table~\ref{tab:problem_dims}, and let $\mathcal{I}_{sub}=\{3,7,10,13,16,23\}$ represent the selected subset, which is designed as a representative microcosm of the full suite, preserving diversity in objective dimensionality, decision space size, and landscape difficulty. Based on this subset, the following two settings are considered:
\begin{itemize}
    \item \textbf{Setting A (3:1).} The meta-policy is trained on $\mathcal{I}_{all} \setminus \mathcal{I}_{sub}$ (18 problems) and evaluated on $\mathcal{I}_{sub}$ (6 problems). This setting examines whether invariant scheduling strategies can be distilled from broad problem exposure.
    \item \textbf{Setting B (1:3):} The meta-policy is trained on $\mathcal{I}_{sub}$ and tested on the remaining 18 problems in $\mathcal{I}_{all} \setminus \mathcal{I}_{sub}$. This setting emphasizes robustness under limited training diversity and generalization to largely unseen domains.
\end{itemize}

\begin{table}[h]
\caption{Parameter settings for the 24 benchmark problems.}
\label{tab:problem_dims}
\centering
\small
\begin{tabular}{lccc|lccc}
\toprule
\textbf{ID} & \textbf{Problem} & \textbf{}{$n_{var}$} & \textbf{$n_{obj}$} & \textbf{ID} & \textbf{Problem} & \textbf{$n_{var}$} & \textbf{$n_{obj}$} \\
\midrule
1  & OMNITEST & 2  & 2 & 13 & DTLZ5 & 30 & 3 \\
2  & KURSAWE  & 3  & 2 & 14 & DTLZ6 & 30 & 3 \\
3  & ZDT1     & 30 & 2 & 15 & DTLZ7 & 30 & 3 \\
4  & ZDT2     & 30 & 2 & 16 & WFG1  & 10 & 3 \\
5  & ZDT3     & 30 & 2 & 17 & WFG2  & 10 & 3 \\
6  & ZDT4     & 10 & 2 & 18 & WFG3  & 10 & 3 \\
7  & ZDT5     & 80 & 2 & 19 & WFG4  & 10 & 5 \\
8  & ZDT6     & 10 & 2 & 20 & WFG5  & 30 & 5 \\
9  & DTLZ1    & 10 & 5 & 21 & WFG6  & 30 & 5 \\
10 & DTLZ2    & 10 & 5 & 22 & WFG7  & 30 & 5 \\
11 & DTLZ3    & 10 & 5 & 23 & WFG8  & 30 & 5 \\
12 & DTLZ4    & 10 & 5 & 24 & WFG9  & 30 & 5 \\
\bottomrule
\end{tabular}
\end{table}
\begin{figure*}[t] 
    \centering
    \begin{subfigure}{\textwidth}
        \centering
        \includegraphics[width=\linewidth]{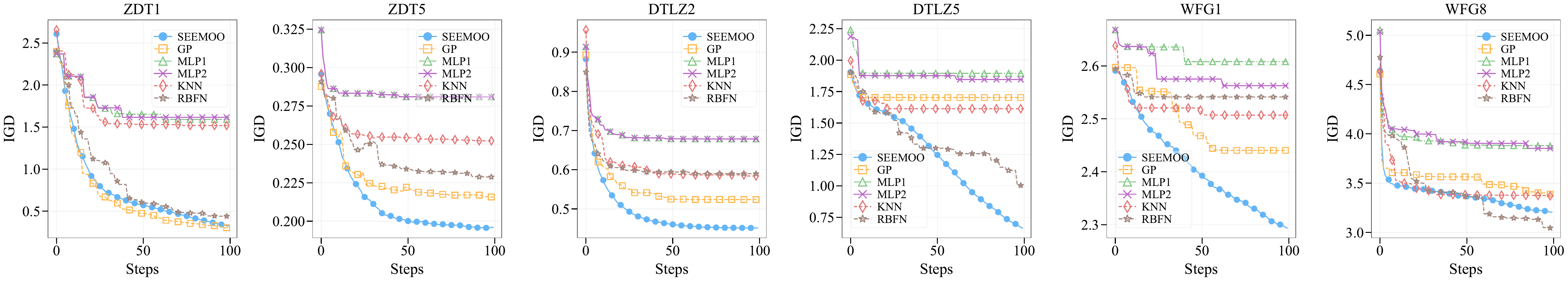}
        \caption{Experimental results for Setting A}
        \Description{}
        \label{fig:igd_ex1}
    \end{subfigure}
    \begin{subfigure}{\textwidth}
        \centering
        \includegraphics[width=\linewidth]{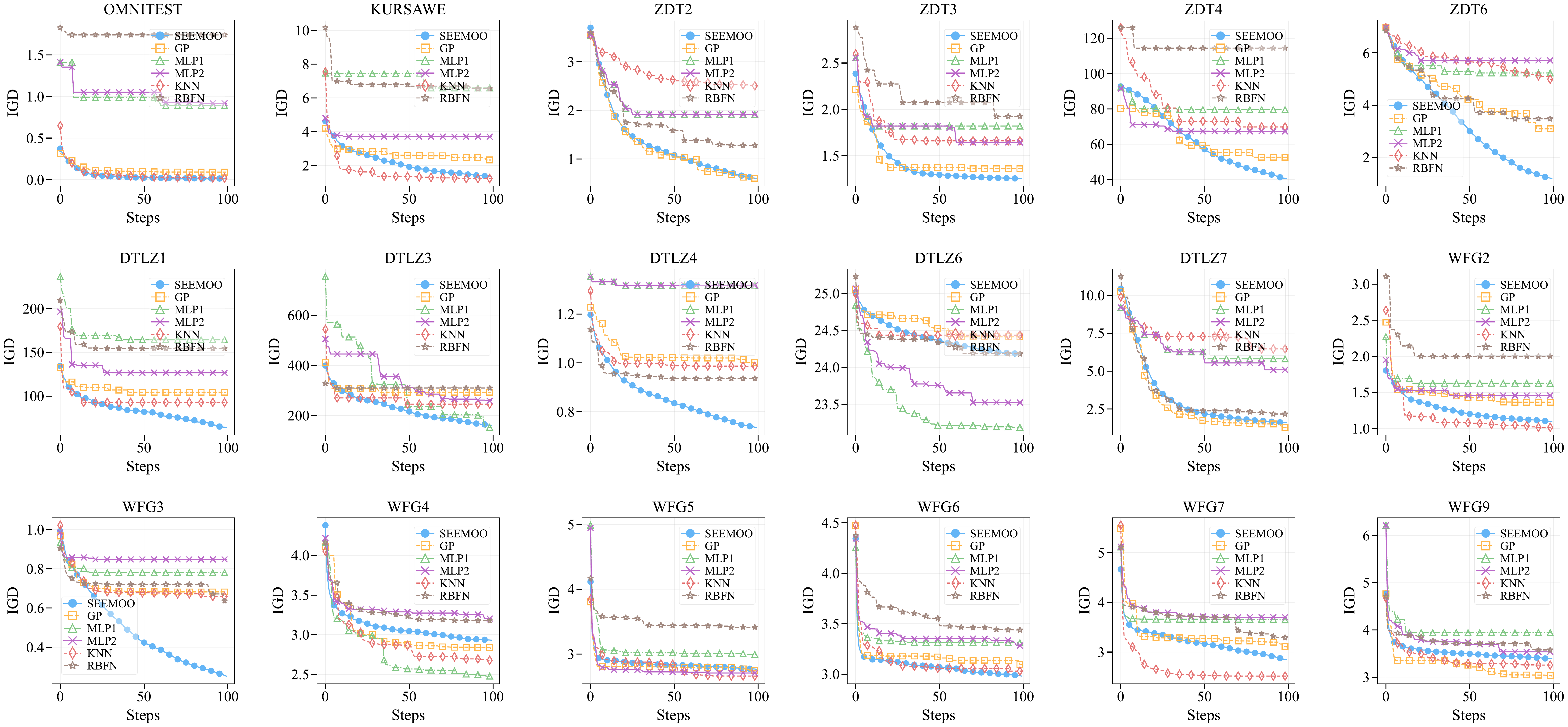}
        \caption{Experimental results for Setting B}
        \label{fig:igd_ex2}
    \end{subfigure}
    \caption{IGD convergence curves of SEEMOO and five baselines across various test problems.}
    \label{fig:total_igd_comparison}
    \Description{The figure presents a comprehensive comparison of IGD convergence curves. 
    Part (a) displays Setting A results, showing several line charts where the x-axis represents optimization steps and the y-axis represents the IGD metric. 
    The SEEMOO curve consistently demonstrates a steeper descent and achieves lower final IGD values compared to five baseline algorithms across different test instances. 
    Part (b) illustrates Setting B with a similar layout, confirming the robustness of SEEMOO as its convergence lines remain below the baselines, indicating superior efficiency in finding the pareto front.}
\end{figure*}
\subsubsection{Baselines}
We compare SEEMOO against five baselines that rely on fixed surrogate models, namely GP~\cite{zhou2005study}, MLP1~\cite{jin2005comprehensive} with low budget, MLP2~\cite{jin2005comprehensive} with high budget, KNN~\cite{he2007hybridisation},  RBFN~\cite{regis2013combining}, to assess the effectiveness of dynamic surrogate scheduling.
\subsubsection{Settings}
For both SEEMOO and all baseline methods, the optimization process follows a fixed evaluation budget of $maxFEs=550$ real function evaluations, reflecting expensive optimization scenarios. During the optimization process, We fix the initial sampling size at $N = 50$. Over $t_{max}=100$ decision steps, a batch of $b=5$ promising candidates is selected for true evaluation per step.

For SEEMOO's specific training details, the DQN agent is trained over 50 epochs on a distribution of training problems. We use the Adam optimizer with a learning rate $\alpha=10^{-3}$ and a discount factor $\gamma=0.9$. An experience replay buffer of size 256 is maintained, with mini-batches of size 32 sampled for each update. To balance exploration and exploitation, an $\epsilon$-greedy strategy is employed, where $\epsilon$ linearly decays from 0.8 to 0.01 during training and remains fixed at 0.01 during testing.

\begin{figure*}[t] 
    \centering
    \begin{subfigure}{\textwidth}
        \centering
        \includegraphics[width=\linewidth]{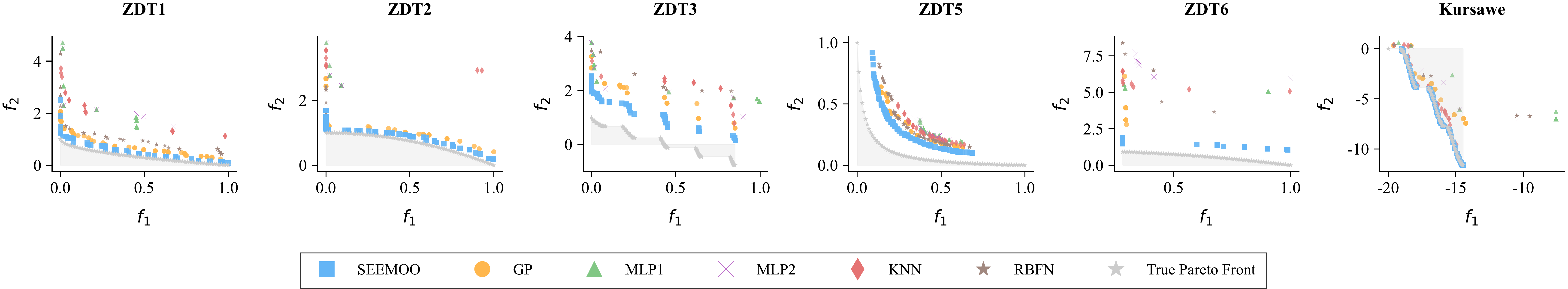}
        \caption{Pareto fronts comparison on two-dimensional problems}
        \label{fig:pf_2d}
    \end{subfigure}\\
    \vspace{5mm}
    \begin{subfigure}{\textwidth}
        \centering
        \includegraphics[width=\linewidth]{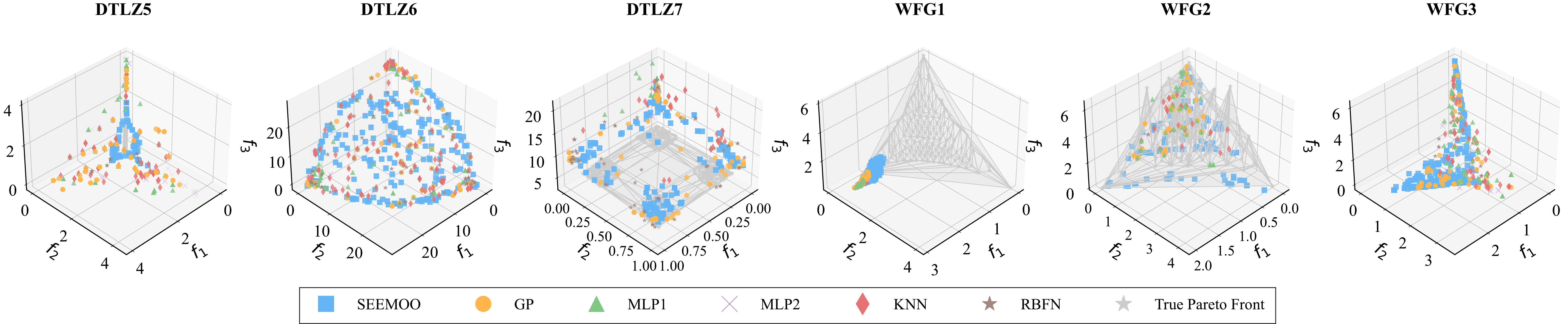}
        \caption{Pareto fronts comparison on three-dimensional problems}
        \label{fig:pf_3d}
    \end{subfigure}
    \caption{Visual comparison of pareto fronts obtained by SEEMOO and baselines.}
    \label{fig:pf_comparison}
    \Description{This figure visualizes the pareto fronts obtained by various algorithms. 
    Part (a) shows two-dimensional objective spaces where the solutions are plotted as scatter points. 
    The points generated by SEEMOO are significantly closer to the true pareto front and exhibit a more uniform distribution than the baselines. 
    Part (b) shows three-dimensional objective spaces with solution sets forming surfaces. 
    SEEMOO's results cover a wider range of the front and maintain better diversity.}
\end{figure*}
\subsection{Comparison Analysis}
To comprehensively evaluate SEEMOO, we conduct analyses from two perspectives to assess its optimization performance and generalization capability. The corresponding IGD convergence curves are illustrated in Figure~\ref{fig:total_igd_comparison}.

\textbf{Performance against baselines.} SEEMOO demonstrates a generally superior performance profile across the ZDT, DTLZ, and WFG problem families. In the ZDT and DTLZ series, it achieves notable IGD reductions on complex multimodal instances (e.g., DTLZ1 and DTLZ3), confirming that dynamic scheduling is particularly effective for approximating pareto fronts. While SEEMOO's advantage narrows on some instances (e.g., DTLZ6) where a specialized static surrogate performs comparably, it consistently avoids the failures that can afflict single‑model baselines. In the more challenging WFG series, characterized by higher dimensionality and complex front geometries, SEEMOO remains competitive, though its advantage is less pronounced, reflecting stable adaptation across diverse landscapes.

\textbf{Generalization across training‑testing partitions.} Furthermore, we examine SEEMOO's generalization performance under two data‑partition scenarios (3:1 and 1:3), corresponding to Figure \ref{fig:igd_ex1} and Figure \ref{fig:igd_ex2}, respectively. The results show that SEEMOO maintains stable performance even in the 1:3 setting. This indicate that the learned meta‑policy captures invariant optimization‑state features rather than memorizing specific problem instances, enabling robust adaption even when training data is limited.

In summary, SEEMOO achieves a balanced trade-off between specialized local modeling and global generalization, which validates the practicality of adopting a meta‑RL approach for dynamic surrogate management in expensive multi‑objective optimization.
\subsection{Ablation Studies}

\begin{table*}[t]
\caption{Comparative results of ablation on surrogate pool and reward design.}
\label{tab:ablation_full}
\centering
\small 
\tabcolsep=2pt 
\resizebox{0.85\textwidth}{!}{
\begin{tabular}{ccccccccccccccc}
\toprule
\textbf{Experiments} & \multicolumn{2}{c}{\textbf{SEEMOO}} & \multicolumn{2}{c}{\textbf{w/o GP}} & \multicolumn{2}{c}{\textbf{w/o KNN}} & \multicolumn{2}{c}{\textbf{w/o MLPs}} & \multicolumn{2}{c}{\textbf{w/o RBFN}} & \multicolumn{2}{c}{\textbf{Binary-Reward}} & \multicolumn{2}{c}{\textbf{HV-Reward}} \\
\midrule
\textbf{Problem} & mean & \multirow{2}{*}{rank} & mean & \multirow{2}{*}{rank} & mean & \multirow{2}{*}{rank} & mean & \multirow{2}{*}{rank} & mean & \multirow{2}{*}{rank} & mean & \multirow{2}{*}{rank} & mean & \multirow{2}{*}{rank} \\
 & (std) & & (std) & & (std) & & (std) & & (std) & & (std) & & (std) & \\
\midrule
\multirow{2}{*}{OMNITEST} & \textbf{4.950e-02} & \textbf{1} & 5.580e-02 & 2 & 6.440e-02 & 6 & 5.820e-02 & 3 & 1.977e-01 & 7 & 5.910e-02 & 5 & 5.890e-02 & 4 \\
                          & ($\pm$6.960e-02) & & ($\pm$9.890e-02) & & ($\pm$1.393e-01) & & ($\pm$1.128e-01) & & ($\pm$1.835e-01) & & ($\pm$9.290e-02) & & ($\pm$9.300e-02) & \\
\midrule
\multirow{2}{*}{KURSAWE}  & 1.778e+00 & 2 & 2.573e+00 & 5 & \textbf{1.697e+00} & \textbf{1} & 2.392e+00 & 3 & 3.897e+00 & 7 & 3.238e+00 & 6 & 2.476e+00 & 4 \\
                          & ($\pm$1.127e+00) & & ($\pm$1.778e+00) & & ($\pm$1.551e+00) & & ($\pm$1.552e+00) & & ($\pm$1.332e+00) & & ($\pm$2.502e+00) & & ($\pm$2.295e+00) & \\
\midrule
\multirow{2}{*}{ZDT2}     & \textbf{1.204e+00} & \textbf{1} & 1.453e+00 & 6 & 1.222e+00 & 2 & 1.334e+00 & 5 & 2.209e+00 & 7 & 1.288e+00 & 3 & 1.288e+00 & 3 \\
                          & ($\pm$7.311e-01) & & ($\pm$6.890e-01) & & ($\pm$7.754e-01) & & ($\pm$7.099e-01) & & ($\pm$6.237e-01) & & ($\pm$7.194e-01) & & ($\pm$7.194e-01) & \\
\midrule
\multirow{2}{*}{ZDT3}     & \textbf{1.235e+00} & \textbf{1} & 1.398e+00 & 5 & 1.335e+00 & 2 & 1.489e+00 & 6 & 1.506e+00 & 7 & 1.371e+00 & 3 & 1.371e+00 & 4 \\
                          & ($\pm$3.034e-01) & & ($\pm$6.235e-01) & & ($\pm$4.972e-01) & & ($\pm$5.233e-01) & & ($\pm$6.806e-01) & & ($\pm$4.994e-01) & & ($\pm$4.994e-01) & \\
\midrule
\multirow{2}{*}{ZDT4}     & 5.681e+01 & 3 & 5.237e+01 & 2 & 6.861e+01 & 7 & 5.999e+01 & 5 & \textbf{4.801e+01} & \textbf{1} & 5.958e+01 & 4 & 6.129e+01 & 6 \\
                          & ($\pm$2.178e+01) & & ($\pm$3.000e+01) & & ($\pm$3.539e+01) & & ($\pm$2.640e+01) & & ($\pm$1.694e+01) & & ($\pm$2.551e+01) & & ($\pm$2.338e+01) & \\
\midrule
\multirow{2}{*}{ZDT6}     & 3.309e+00 & 6 & 2.954e+00 & 2 & 3.176e+00 & 5 & \textbf{2.853e+00} & \textbf{1} & 3.552e+00 & 7 & 2.985e+00 & 3 & 2.985e+00 & 3 \\
                          & ($\pm$1.800e+00) & & ($\pm$1.727e+00) & & ($\pm$1.656e+00) & & ($\pm$1.764e+00) & & ($\pm$1.346e+00) & & ($\pm$1.754e+00) & & ($\pm$1.754e+00) & \\
\midrule
\multirow{2}{*}{DTLZ1}    & 8.723e+01 & 6 & 7.709e+01 & 2 & \textbf{7.479e+01} & \textbf{1} & 8.194e+01 & 5 & 1.128e+02 & 7 & 7.858e+01 & 3 & 7.858e+01 & 3 \\
                          & ($\pm$2.443e+01) & & ($\pm$2.630e+01) & & ($\pm$4.332e+01) & & ($\pm$3.244e+01) & & ($\pm$3.042e+01) & & ($\pm$2.973e+01) & & ($\pm$2.973e+01) & \\
\midrule
\multirow{2}{*}{DTLZ3}    & 2.314e+02 & 5 & 2.270e+02 & 4 & \textbf{1.868e+02} & \textbf{1} & 2.215e+02 & 3 & 2.071e+02 & 2 & 2.463e+02 & 6 & 2.463e+02 & 6 \\
                          & ($\pm$7.335e+01) & & ($\pm$5.933e+01) & & ($\pm$8.265e+01) & & ($\pm$7.264e+01) & & ($\pm$1.093e+02) & & ($\pm$7.077e+01) & & ($\pm$7.077e+01) & \\
\midrule
\multirow{2}{*}{DTLZ4}    & \textbf{8.653e-01} & \textbf{1} & 9.824e-01 & 4 & 1.052e+00 & 6 & 8.852e-01 & 2 & 1.051e+00 & 6 & 9.526e-01 & 3 & 1.019e+00 & 5 \\
                          & ($\pm$1.526e-01) & & ($\pm$1.875e-01) & & ($\pm$1.862e-01) & & ($\pm$1.894e-01) & & ($\pm$1.613e-01) & & ($\pm$1.842e-01) & & ($\pm$1.335e-01) & \\
\midrule
\multirow{2}{*}{DTLZ6}    & 2.416e+01 & 3 & 2.421e+01 & 4 & 2.424e+01 & 5 & 2.397e+01 & 2 & \textbf{2.325e+01} & \textbf{1} & 2.444e+01 & 6 & 2.444e+01 & 6 \\
                          & ($\pm$2.960e-01) & & ($\pm$3.803e-01) & & ($\pm$3.166e-01) & & ($\pm$4.113e-01) & & ($\pm$7.018e-01) & & ($\pm$3.664e-01) & & ($\pm$3.664e-01) & \\
\midrule
\multirow{2}{*}{DTLZ7}    & \textbf{3.182e+00} & \textbf{1} & 3.563e+00 & 4 & 3.394e+00 & 2 & 3.454e+00 & 3 & 5.799e+00 & 7 & 3.943e+00 & 5 & 3.943e+00 & 5 \\
                          & ($\pm$2.363e+00) & & ($\pm$2.538e+00) & & ($\pm$2.394e+00) & & ($\pm$2.437e+00) & & ($\pm$2.119e+00) & & ($\pm$2.322e+00) & & ($\pm$2.322e+00) & \\
\midrule
\multirow{2}{*}{WFG2}     & 1.291e+00 & 2 & 1.919e+00 & 6 & 1.843e+00 & 5 & 2.166e+00 & 7 & \textbf{1.290e+00} & \textbf{1} & 1.779e+00 & 4 & 1.758e+00 & 3 \\
                          & ($\pm$2.712e-01) & & ($\pm$5.795e-01) & & ($\pm$3.889e-01) & & ($\pm$5.481e-01) & & ($\pm$3.649e-01) & & ($\pm$6.992e-01) & & ($\pm$6.986e-01) & \\
\midrule
\multirow{2}{*}{WFG3}     & 5.292e-01 & 6 & 4.033e-01 & 4 & \textbf{3.879e-01} & \textbf{1} & 4.029e-01 & 3 & 5.352e-01 & 7 & 3.898e-01 & 2 & 4.156e-01 & 5 \\
                          & ($\pm$2.367e-01) & & ($\pm$2.499e-01) & & ($\pm$2.061e-01) & & ($\pm$2.723e-01) & & ($\pm$1.513e-01) & & ($\pm$2.576e-01) & & ($\pm$2.715e-01) & \\
\midrule
\multirow{2}{*}{WFG4}     & 3.167e+00 & 2 & 3.990e+00 & 6 & 3.623e+00 & 4 & 3.577e+00 & 3 & \textbf{2.850e+00} & \textbf{1} & 3.812e+00 & 5 & 3.794e+00 & 5 \\
                          & ($\pm$3.511e-01) & & ($\pm$6.606e-01) & & ($\pm$5.316e-01) & & ($\pm$7.138e-01) & & ($\pm$7.291e-01) & & ($\pm$6.646e-01) & & ($\pm$6.609e-01) & \\
\midrule
\multirow{2}{*}{WFG5}     & 2.935e+00 & 2 & 3.920e+00 & 7 & 3.585e+00 & 3 & 3.709e+00 & 4 & \textbf{2.358e+00} & \textbf{1} & 3.820e+00 & 5 & 3.869e+00 & 6 \\
                          & ($\pm$2.375e-01) & & ($\pm$6.786e-01) & & ($\pm$5.385e-01) & & ($\pm$5.054e-01) & & ($\pm$5.834e-01) & & ($\pm$6.733e-01) & & ($\pm$7.059e-01) & \\
\midrule
\multirow{2}{*}{WFG6}     & 3.089e+00 & 2 & 3.736e+00 & 3 & 4.027e+00 & 5 & 4.007e+00 & 4 & \textbf{2.563e+00} & \textbf{1} & 4.122e+00 & 6 & 4.257e+00 & 7 \\
                          & ($\pm$2.606e-01) & & ($\pm$6.727e-01) & & ($\pm$3.563e-01) & & ($\pm$5.205e-01) & & ($\pm$6.278e-01) & & ($\pm$4.810e-01) & & ($\pm$4.993e-01) & \\
\midrule
\multirow{2}{*}{WFG7}     & 3.083e+00 & 2 & 3.683e+00 & 7 & 3.403e+00 & 3 & 3.481e+00 & 4 & \textbf{2.726e+00} & \textbf{1} & 3.627e+00 & 6 & 3.516e+00 & 5 \\
                          & ($\pm$4.105e-01) & & ($\pm$9.230e-01) & & ($\pm$6.535e-01) & & ($\pm$1.159e+00) & & ($\pm$5.220e-01) & & ($\pm$8.439e-01) & & ($\pm$6.884e-01) & \\
\midrule
\multirow{2}{*}{WFG9}     & 3.536e+00 & 2 & 4.457e+00 & 7 & 4.189e+00 & 5 & 4.325e+00 & 6 & \textbf{2.822e+00} & \textbf{1} & 4.174e+00 & 4 & 4.021e+00 & 3 \\
                          & ($\pm$3.678e-01) & & ($\pm$6.354e-01) & & ($\pm$6.112e-01) & & ($\pm$8.287e-01) & & ($\pm$4.422e-01) & & ($\pm$5.396e-01) & & ($\pm$6.152e-01) & \\
\midrule
\textbf{Avg Rank} & \multicolumn{2}{c}{\textbf{2.67}} & \multicolumn{2}{c}{4.5} & \multicolumn{2}{c}{3.61} & \multicolumn{2}{c}{3.83} & \multicolumn{2}{c}{4.00} & \multicolumn{2}{c}{4.44} & \multicolumn{2}{c}{4.56} \\
\bottomrule
\end{tabular}
}
\end{table*}
To systematically evaluate the contribution of each component and justify the design choices in SEEMOO, we conduct ablation studies across the benchmark suite. We derive six variants, categorized by the ablated component: surrogate model selection and reward function design.
\subsubsection{Surrogate Model Selection} To investigate the necessity and contribution of each candidate surrogate in the model pool $\mathcal{M}$, we remove specific model architectures and analyze the subsequent degradation in optimization performance. The four ablated variants are defined as follows:
\begin{itemize}
    \item \textbf{w/o GP}: Removes GP~\cite{zhou2005study}, testing the impact of losing uncertainty quantification and high-precision local modeling.
    \item \textbf{w/o MLPs}: Removes both MLP1~\cite{jin2005comprehensive} and MLP2~\cite{jin2005comprehensive}, assessing the loss of global feature extraction capability.
    \item \textbf{w/o KNN}: Removes KNN~\cite{he2007hybridisation}, evaluating robustness in capturing nonlinear local structures.
    \item \textbf{w/o RBFN}: Removes RBFN~\cite{regis2013combining}, which provides essential smooth interpolation for continuous fitness landscapes.
\end{itemize}
\textbf{Results Analysis}: In Table 1, the SEEMOO framework achieves the superior Average Rank (2.67). The most significant performance drop occurs in w/o GP (Avg Rank 4.50), confirming that GP is the cornerstone for high-precision modeling in evaluation-constrained environments. While w/o KNN occasionally excels on problems with irregular fronts, the full ensemble ensures the best stability.

\subsubsection{Reward Function Design}
The reinforcement learning agent in SEEMOO is guided by a reward signal that reflects the improvement in the quality of the pareto front approximation. To justify our choice of IGD‑based reward, we compare it against two common alternative reward formulations:

1) \emph{Binary-Reward}. The reward $r_t$ is simplified into a binary signal, defined as:
\begin{equation}
    r_t = \begin{cases} 
            1, & \text{if } IGD_t < IGD_{t-1}\\ 
            0, & \text{otherwise} 
          \end{cases}
\end{equation}
This variant tests whether sparse feedback is sufficient for the DQN agent to learn effective scheduling policies.

2) \emph{HV-Reward}. IGD is replaced by Hypervolume (HV), which evaluates the sensitivity of the meta-strategy to different performance indicators. Since larger HV values indicate better performance, the reward is defined as: 
\begin{equation}
    r_t = \frac{HV_t - HV_{t-1}}{HV_{max} - HV_{min}}
\end{equation}
where $HV_{t}$ and $HV_{t-1}$ denote the HV values at the current and previous optimization steps. $HV_{max}$ and $HV_{min}$ denote the maximum and minimum HV value observed globally across all decision steps up to step $t$.

\textbf{Results Analysis}: Table 1 reveals that the Binary reward suffers from sparse feedback (Avg Rank 4.44), while the HV-based reward (Avg Rank 4.56) fails to provide a sufficiently informative gradient for meta-learning. In contrast, our IGD-based reward optimally balances between stability and efficiency.
\subsubsection{Concluding Remarks on Design}
The ablation studies confirm the internal consistency and necessity of the key design choices in SEEMOO. Two main conclusions can be drawn:

1) Removing any surrogate from the model pool consistently degrades optimization performance, indicating that each surrogate provides complementary information to the learned scheduling policy. Overall, these results confirm that maintaining a heterogeneous surrogate pool is critical, and that no single surrogate architecture is sufficient on its own.

2) The reward formulation plays a critical role in policy learning. Sparse or discontinuous feedback leads to inferior performance, indicating that it is insufficient for guiding effective surrogate scheduling. In contrast, dense and smoothly varying reward design based on IGD provide more reliable learning signals, resulting in more stable convergence and improved generalization across problems.
\section{Conclusion}
This paper presents SEEMOO, a reinforcement learning-assisted surrogate ensemble framework for expensive multi-objective optimization. By replacing manual model selection with a learning-driven scheduling policy, SEEMOO effectively mitigates designer bias and enhances adaptability across diverse landscapes, which integrates an attention-based state extractor for universal problem representation and a DQN agent to adaptively select optimal surrogates from a heterogeneous pool.

Experimental results show that SEEMOO consistently outperforms single-surrogate baselines across various benchmark problems. Notably, SEEMOO demonstrates strong generalization capability, consistently achieving competitive performance on unseen problem instances. These findings validate the effectiveness of modeling surrogate management as a meta-RL task, providing a promising direction to overcome the limitations of static ensemble designs in expensive multi-objective optimization. Future work will focus on: 1) expanding the surrogate pool with aditional model types, 2) integrating SEEMOO with various MOEA architectures, 3) validating the framework's practical utility and generalization capability on real-world engineering optimization tasks.

\bibliographystyle{ACM-Reference-Format} 
\bibliography{sample-base}
\end{document}